\documentclass[letterpaper, 10 pt, conference]{ieeeconf}  

\IEEEoverridecommandlockouts                              

\usepackage{graphics} 
\usepackage{epsfig} 
\usepackage{mathptmx} 
\usepackage{times} 
\usepackage{amsmath} 
\usepackage{amssymb}  
\usepackage{cite}
\usepackage{multirow}
\usepackage{subcaption}
\usepackage{booktabs}
\usepackage{cite}

\title{\LARGE \bf
 Gaze-Guided 3D Hand Motion Prediction for Detecting Intent in Egocentric Grasping Tasks
}

\author{Yufei He$^{1}$, Xucong Zhang$^{2}$, and Arno H. A. Stienen$^{1}$
\thanks{$^{1}$ Department of Biomechanical Engineering, Faculty of Mechanical Engineering, Delft University of Technology}%
\thanks{$^{2}$ Department of Intelligent Systems, Delft University of Technology}%
}

\begin{document}

\maketitle

\thispagestyle{empty}
\pagestyle{empty}

\begin{abstract}
Human intention detection with hand motion prediction is critical to drive the upper-extremity assistive robots in neurorehabilitation applications. 
However, the traditional methods relying on physiological signal measurement are restrictive and often lack environmental context. 
We propose a novel approach that predicts future sequences of both hand poses and joint positions.
This method integrates gaze information, historical hand motion sequences, and environmental object data, adapting dynamically to the assistive needs of the patient without prior knowledge of the intended object for grasping. 
Specifically, we use a vector-quantized variational autoencoder for robust hand pose encoding with an autoregressive generative transformer for effective hand motion sequence prediction. We demonstrate the usability of these novel techniques in a pilot study with healthy subjects.
To train and evaluate the proposed method, we collect a dataset consisting of various types of grasp actions on different objects from multiple subjects. 
Through extensive experiments, we demonstrate that the proposed method can successfully predict sequential hand movement. Especially, the gaze information shows significant enhancements in prediction capabilities, particularly with fewer input frames, highlighting the potential of the proposed method for real-world applications.
\end{abstract}

\section{INTRODUCTION}

\begin{figure*}[htbp]
\centering
\includegraphics[width=0.9\textwidth]{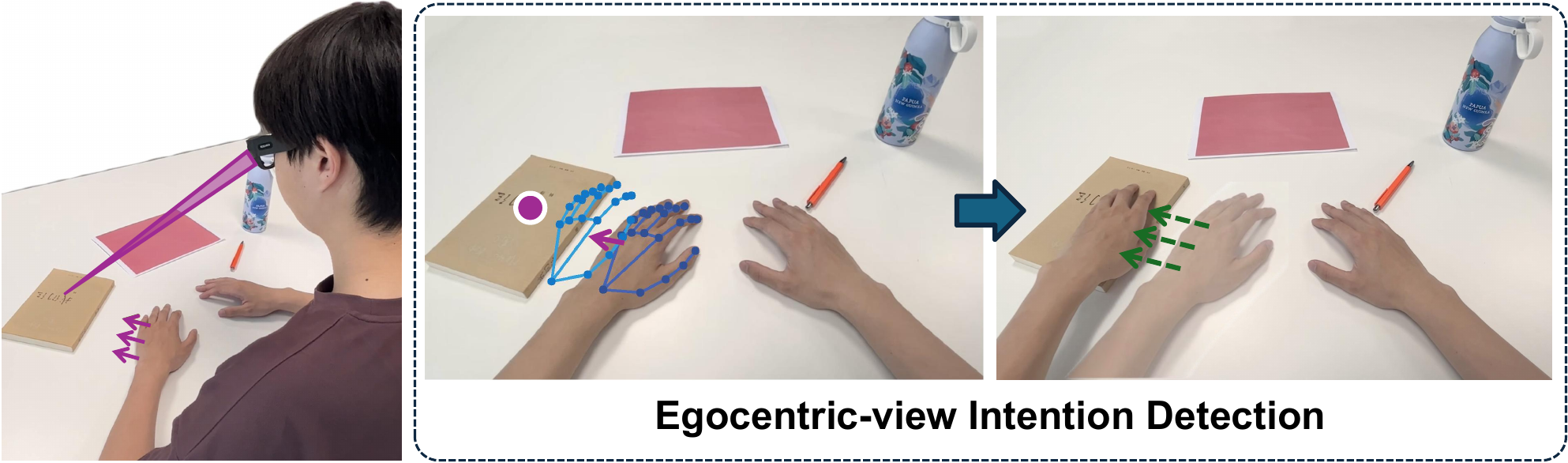}
\caption{
\textbf{Overview of gaze-guided human intention detection.} Left: The user wears eye-tracking glasses to capture gaze fixation points (purple ray), initial hand motion (purple arrow), and object locations as input. Right: Using egocentric-view data, including eye fixation points (purple dot) and hand motion (blue hand skeletons), the system predicts a sequence of hand motions leading to the final grasping action (green arrow). 
}
\label{fig:teaser}
\end{figure*}

Upper extremity movement disorders caused by stroke or traumatic brain injury often limit daily task performance~\cite{newengland_upperrehab,luo2024research}, driving the development of assistive robots that enhance re-acquisition of motor function through adaptive exercises~\cite{basteris2014training}. 
Intention detection is crucial for robots to assist patients, as it allows the robot to understand the desired actions of the user and provide customized assistance \cite{wang2023sensors}. The transition from clinic-based rehabilitation robots to home assistive robots that support everyday tasks highlights the importance of quickly understanding and assisting with user intentions in domestic settings. Conventional methods such as surface electromyography (sEMG) and electroencephalography (EEG) have been used for intention detection, which can directly measure physiological signals \cite{gu2022review}. However, these methods usually restrict movement, require frequent re-calibrations, and lack the perception of environmental context. Additionally, these signals in post-stroke conditions are also disturbed \cite{ag2023qualitative} and thus become hard to correlate with movement. 

Recent developments in computer vision have enhanced the utility of vision signals for intention interpretation. This progress enables robots to learn from natural human behaviors and observe environmental interactions, thus significantly improving their ability to assist in a manner that aligns closely with human needs \cite{guo2021human}. 
A variety of datasets capturing both hand and object interactions have been developed \cite{taheri2020grab,hampali2020honnotate,chao2021dexycb, kwon2021h2o,yang2022oakink,liu2022hoi4d,fan2023arctic}, facilitating studies into grasp generation \cite{jiang2021hand,karunratanakul2020grasping,karunratanakul2021skeleton}. For example, Jiang et al. \cite{jiang2021hand} refined grasping gestures using contact maps on objects. However, these studies have been limited to generating static gestures. 

Gaze information, which reveals user intent by identifying focus areas before physical actions, enhances human motion prediction for assistive robots\cite{Lin2023NeuralCO,crocher2021towards}. Nevertheless, the application of gaze data often focuses on classification tasks and may suffer from inaccuracies due to false positives \cite{guo2021human}. 
Although research has utilized egocentric signals \cite{girdhar2021anticipative,turnoff,lstm_action_anti}, or combined them with gaze data \cite{li2021eye,adebayo2022hand}, the focus has generally been on semantic predictions while the capability of explicit future hand motion prediction is missing. Developing this capability is essential for assistive robots to offer effective assistance throughout training.
Studies for hand motion prediction often depend on explicit conditions such as the geometry or position of the object \cite{ghosh2023imos,li2024task,taheri2022goal,wu2022saga,christen2022d,zheng2023cams}, initial or final hand positions \cite{ghosh2023imos,li2024task,taheri2022goal,wu2022saga,christen2022d,zheng2023cams}, trajectories \cite{zhang2021manipnet}, or textual descriptions \cite{ghosh2023imos}. For example, Christen et al. \cite{christen2022d} introduced a method to synthesize diverse hand movements based on the start and end poses of an object.

In response to this challenge, we propose a novel task for intention detection. Given a set of potential grasping objects and initial hand movements, we want to predict future intended hand motions, including hand poses and joint positions. 
This task focuses on two fundamental aspects: 1) utilizing only implicit environmental context, and 2) producing explicit hand motion outputs represented by 21 hand key points. To tackle this task, we have developed a method that employs gaze- and egocentric-view visual signals to predict future hand motions. This setting is practical for assistive robot applications because the user can operate the robot with the head-mounted device. 

Specifically for hand motion prediction, we develop a method consisting of a Vector-Quantized Variational AutoEncoder (VQ-VAE) and an auto-regressive generative transformer. The VQ-VAE encodes hand poses to capture in-distribution features for precise motion prediction, while the transformer generates future hand motion sequences from any input frame. Additionally, our feature fusion architecture, consisting of linear layers, integrates gaze and object features to enhance accuracy. 

We validate the generalizability of our model in our self-collected dataset across different subjects and motions, and we explore the impact of various gaze fusion methods on model performance. Our findings indicate robust generalization across diverse validation settings, particularly in distance accuracy. The gaze-integrated model significantly outperforms the no-gaze model, especially with fewer input frames, highlighting the value of gaze information when historical data is limited. 
The results show that, compared with the no-gaze model, our model has the potential to provide accurate and timely predictions in real-time situations.

In summary, our paper introduces a novel approach to hand motion prediction that enhances hand movements in interactive tasks. The main contributions are:

\begin{enumerate}
    \item We propose a new task of hand motion sequence prediction with the goal of driving upper extremity assistive robots.
    
    \item We combine gaze data with egocentric visual signals for hand motion prediction.


    \item We validate that our model generalizes effectively to grasping behaviors, illustrating its broad applicability.

\end{enumerate}



\begin{figure*}
\centering
\subcaptionbox{Hand Motion VQ-VAE}{\includegraphics[width=0.40\textwidth]{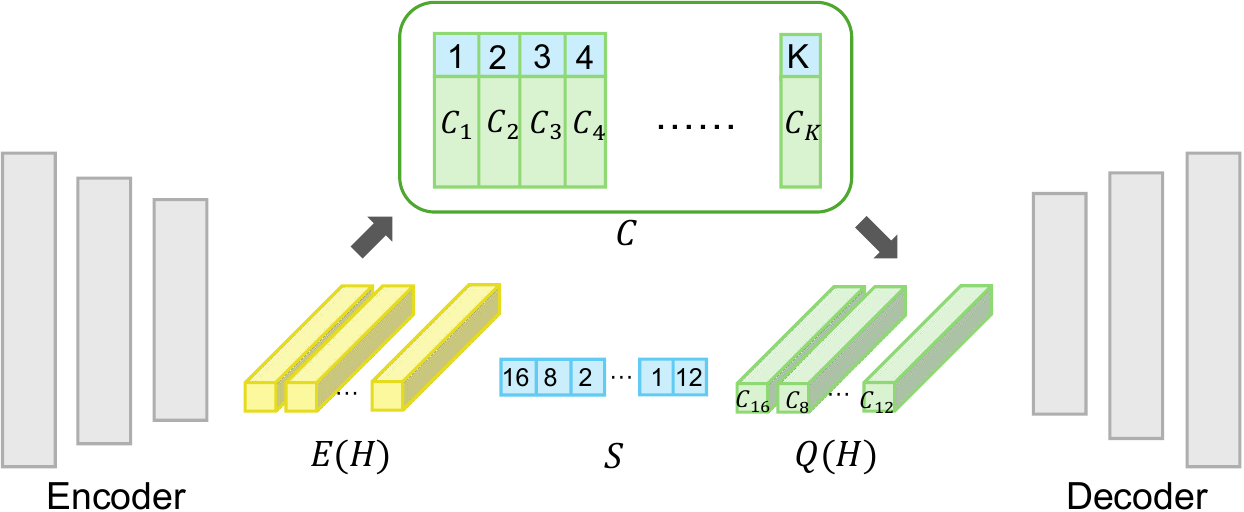}}%
\hfill
\subcaptionbox{Hand Motion Generator}{\includegraphics[width=0.55\textwidth]{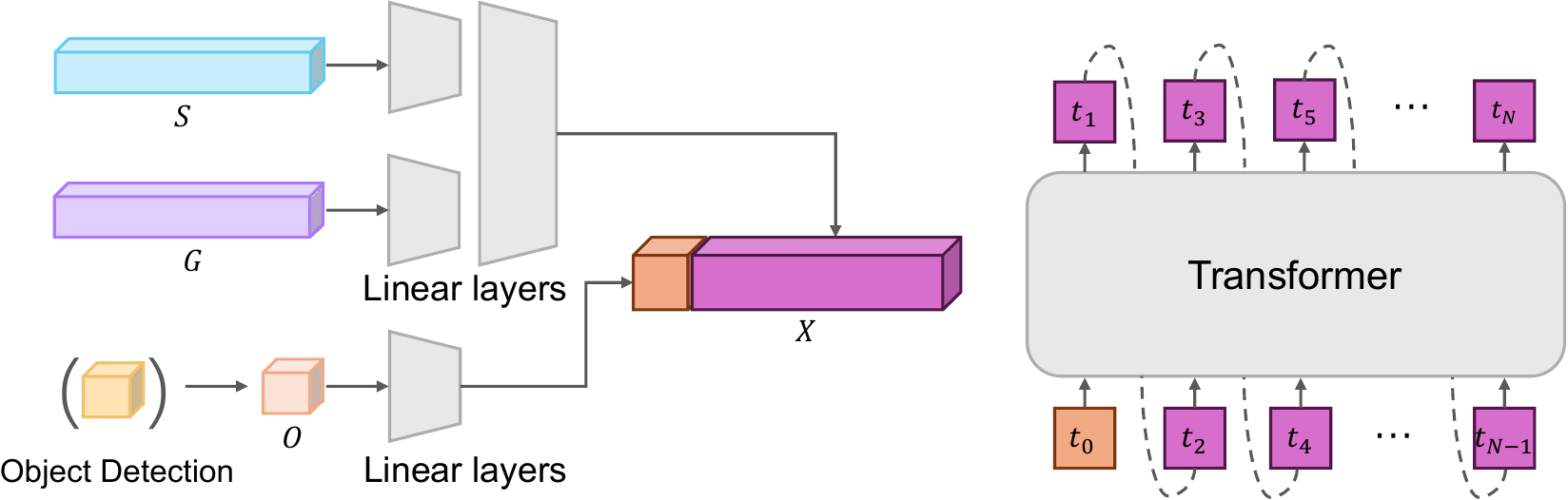}}%
\caption{
\textbf{Overview of our framework for hand motion prediction.} 
The proposed method consists of two main components: (a) Hand-Motion VQ-VAE, which encodes hand motion into codebook $C$ with indices $S$; and (b) Hand Motion Generator, which contains feature fusion layers and a transformer. In feature fusion layers, the encoded hand motion $S$ is integrated with eye-gaze and object features $G$ and $O$, together forming fused feature $X$. The transformer predicts future hand motion indices in an auto-regressive manner using a transformer architecture. These indices are subsequently decoded using the VQ-VAE decoder to obtain the predicted hand motions.}
\label{method:main}
\end{figure*}

\section{METHOD}
Our method performs intention detection via hand motion prediction from the user eye gaze, historical hand motions, and object information.
To perform the task, our method includes two modules: the hand motion VQ-VAE for discrete hand pose codebook learning and the hand motion generator to predict the sequence of hand motions. 

\subsection{Problem Formulation}
The intention detection can be formulated as a model \( M \) that predicts a sequence of future hand motion \( \hat{H} \) based on an initial sequence of hand motion \( H \), a corresponding sequence of eye gazes \( G \), and the representation of the possible interactive objects \( O \) in the first frame. The task is formally defined as:
\begin{equation}
    \hat{H} = M(H, G, O).
\end{equation}
%

\( H = \{h_t\}_{t=1}^{\tau} \) is a sequence of input hand motion, where $\tau$ is the input frame number. The hand pose at frame $t$, denoted by $h_t \in \mathbb{R}^{126}$, is defined by the positions of 21 3D hand joint positions $(x, y, z)$ for both hands. This configuration includes 20 finger joints and one wrist position per hand, according to the Mediapipe framework \cite{lugaresi2019mediapipe}, resulting in a total dimension of $126 = 21 \times 2 \times 3$. 
Similarly, \( G = \{g_t\}_{t=1}^{\tau} \) is a sequence of eye gaze fixation points, represented as $g_t \in \mathbb{R}^3$, is characterized by the 3D eye fixation point $(x, y, z)$ in the world coordinate system. 
Objects in the scene are represented by maximum of four 3D points \( O = \{o^k\} \), where $o_k \in \mathbb{R}^{12}$. 
For instance, a sheet of paper is described using the positions of its four corners, while a pen is represented by the positions of its tip and bottom, reflecting their distinct shapes. The predicted sequence hand motion \(\hat{H} = \{\hat{h}_t\}_{t=\tau+1}^{T} \) consists of hand pose at frame t, $\hat{h}_t \in \mathbb{R}^{126}$, maintains the same dimension as the input hand pose $h_t$ 
starting from frame $\tau+1$ until the end frame $T$.

\subsection{Hand Motion VQ-VAE}
Hand poses have a large space of movements that is difficult to model. A similar problem exists in the human body pose modeling, where the VQ-VAE \cite{van2017vqvae} has been proposed to encode the continuous body movements into discrete classes within a latent space \cite{yasar2023vader}.
We utilize the VQ-VAE to learn multiple hand poses, which can be represented as discrete classes in the motion generation phase. 
An overview of the Hand Motion VQ-VAE model is presented in Fig. \ref{method:main} (a). The encoded features serve as inputs for the hand motion generation network. The codebook is defined as \( C = \{c_i\}_{i=1}^K \), where each \( c_i \) belongs to \( \mathbb{R}^{D_c} \), \( K \) represents the size of the discrete latent space, and \( D_c \) is the dimensionality of each embedding vector. The sequence is encoded as \( E(H) = \{e_t\}_{t=1}^{\lfloor T/l \rfloor} \), with each embedding \( e \in \mathbb{R}^{D_c} \) and \( l \) denoting the downsampling scale. 
The discrete embeddings \( Q=\{q_t\}_{t=1}^{\lfloor T/l \rfloor} \) and indices \( S=\{s_t\}_{t=1}^{\lfloor T/l \rfloor} \) for each frame are computed as:
\begin{gather}
    q_t = \arg\min_{c_i \in C} \|e_t - c_i\|_2 \\
    s_t = \arg\min_{i} \|e_t - c_i\|_2
\end{gather}


\subsubsection{Network Architecture}

Inspired by previous work \cite{zhang2023t2m}, 
the encoder utilizes two convolutional layers with a stride of two for temporal downsampling, reducing the temporal length by a factor of four. This approach not only minimizes computational demands but also reduces noise within the input data. In contrast, the decoder employs nearest-neighbor interpolation for upsampling, facilitating the reconstruction of the complete hand motion sequence.

\subsubsection{Optimization Strategy}
To optimize the VQ-VAE model, the loss function \( L_{vq} \) consists of reconstruction loss, embedding loss, and commitment loss, detailed as follows:

\begin{align}
    &L_{recon} =  
    \begin{cases} 
    0.5 (h_t - \hat{h_t})^2/\beta, & \text{if } |h_t - \hat{h_t}| < \beta \\
    |h_t - \hat{h_t}| - 0.5\beta , & \text{otherwise}
    \end{cases} \\
    &L_{embed} = \|\text{sg}[e_t] - q_t\|_2^2 \\
    &L_{commit} = \gamma \|e_t - \text{sg}[q_t]\|_2^2
\end{align}
where $\beta$ and $\gamma$ are hyper-parameters that influence reconstruction loss and commitment loss, respectively. ``sg'' represents the stop-gradient operator, which prevents the back-propagation of gradient, treating the variable as a constant during the optimization process. The total loss is written as $L_{vq} = L_{recon} + L_{embed} + L_{commit}$.

\subsection{Hand Motion Generator}
With a trained hand-motion VQ-VAE model, the input hand-motion sequence \( H = \{h_t\}_{t=1}^{T_g} \) is encoded into a sequence of quantized indices \( S = \{s_t\}_{t=1}^{T_d}\), where ${T_d} = {\lfloor T_g/l \rfloor}$. As demonstrated in Fig. \ref{method:main} (b), these indices are fused with gaze features \( G \) and conditioned on object features \( O \), serving as inputs \(X\) to the hand-motion generator. This generator operates in an autoregressive manner, producing predicted hand motion indices \( \hat{S} = \{\hat{s}_t\}_{t=1}^{{T_d}+1} \). Given the combined features up to the previous frame \( X_{t-1} \), the probability of each code book index being selected for the next frame hand-motion is calculated as \( p_i(s_t | X_{t-1}) \). The next-frame hand-motion index is determined by:
\begin{equation}
    \hat{s}_t = \arg\max_{i} p_i(s_t | X_{t-1})
\end{equation}

The sequence of indices is mapped to the learned codebook embeddings, forming \( \hat{Q} = \{\hat{q}_t\}_{t=1}^{{T_d}+1} \), where each \( \hat{q}_t \) corresponds to \( c_{\hat{s}_t} \) from the codebook. This encoded sequence \( \hat{Q} \) is then processed by the decoder \( D \), which reconstructs the predicted hand motion sequence \( \hat{H} = \{\hat{h}_t\}_{t=1}^{T_p} \).

\subsubsection{Feature Combination}
To align the dimensional differences between gaze features and hand-motion token embeddings, the gaze features are expanded using a linear layer, producing \( G'=\{g'_t\}_{t=1}^{T_d} \) where each \( g'_t \in D_g \). These embeddings are then concatenated and passed through a linear feature fusion layer followed by a ReLU function, resulting in the combined hand-eye embeddings \( F(S, G)=\{f_t\}_{t=1}^{T_d} \), where each \( f_t \in \mathbb{R}^{D_x} \). 
Notably, we do not apply off-the-shelf object detection here, due to the existence of well-established methods for accurate real-time object detection, allowing us to focus on other aspects of our study. Instead, we manually extract object positions from the first frame and transform them via a linear layer to match \( D_x \), forming \( O' \in \mathbb{R}^{D_x} \), which acts as a conditioning input. The object embeddings are concatenated at the start of the sequence to create \( X = \text{Concat}(O', F) = \{x_t\}_{t=0}^{T_d} \), with each \( x_t \in \mathbb{R}^{D_x} \).

\subsubsection{Decoder-only Transformer Architecture}

We employ decoder-only transformers with masked self-attention layers similar to \cite{zhang2023t2m} for human pose generation, enabling the model to learn input tokens sequentially. 

The masked self-attention is calculated as follows:
\begin{align}
   &Q = XW^Q; K = XW^K; V = XW^V \\
    &Att(Q, K, V) = \text{softmax}\left(\frac{QK^T - M}{\sqrt{D_x}}\right)V  \\
    &M_{i,j} = 
    \begin{cases} 
    0 & \text{if } i \geq j, \\
    -\infty & \text{if } i < j.
    \end{cases}
\end{align}
$W^Q$, $W^K$, and $W^V \in \mathbb{R}^{D_x \times D_x}$ represent the linear projection weights for queries, keys, and values, respectively. $Att$ is the soft attention operation. \( M \) is the mask ensuring predictions for a position do not depend on the following positions.

\subsubsection{Optimization Strategy}
The loss for the transformer model is computed as a classification task as follows:
\begin{equation}
    L_{transformer} = -\sum_{t=1}^{N} w_t \cdot \log(p(\hat{s_t}|X_{t-1})), 
\end{equation}
where \( w_t \) is the weight assigned to each index, and \( N \) is the length of the learned sequence. Specifically, \( w_N \), the weight for the last index, is greater than the weights assigned to other indices to enhance the final hand pose prediction.

\section{EXPERIMENTS AND RESULTS}

\subsection{Dataset Collection}
To evaluate the developed method, we utilize the Project Aria Glasses from Meta \cite{somasundaram2023projectaria} to capture eye-tracking data and an egocentric view of grasping procedures. For this study, 15 volunteers are recruited to participate in the data collection process. Prior to the experiments, participants receive comprehensive instructions detailing the tasks and procedures, and each object is associated with a specific grasping type \cite{grasp}, as shown in Table \ref{table:motions}. The dataset collection is approved by our ethical committee.

\begin{table}[!t]
\renewcommand{\arraystretch}{1.3} 
\caption{Summary of Motions and Interactions}
\label{table_example}
\centering
\small
\resizebox{\columnwidth}{!}{%
\begin{tabular}{l|ccc}
\hline
Motion                & Grasping Type \cite{grasp} & Involved Object & Num. Hands \\
\hline
Pick up a bottle      & Type A        & Bottle          & 1          \\
Move a piece of paper & Type B        & Paper           & 1          \\
Pick up a book        & Type C        & Book            & 1          \\
Pick up a phone       & Type C        & Phone           & 1          \\
Pick up a pen         & Type D        & Pen             & 1          \\
Pick (an) earphone(s) & Type D        & Earphone(s)     & 1 or 2     \\
Write on paper        & Type B, D     & Paper, Pen      & 2          \\
\hline
\end{tabular}
}
\label{table:motions}
\end{table}

\begin{figure}[t!]
\centering
\includegraphics[width=0.46\textwidth]{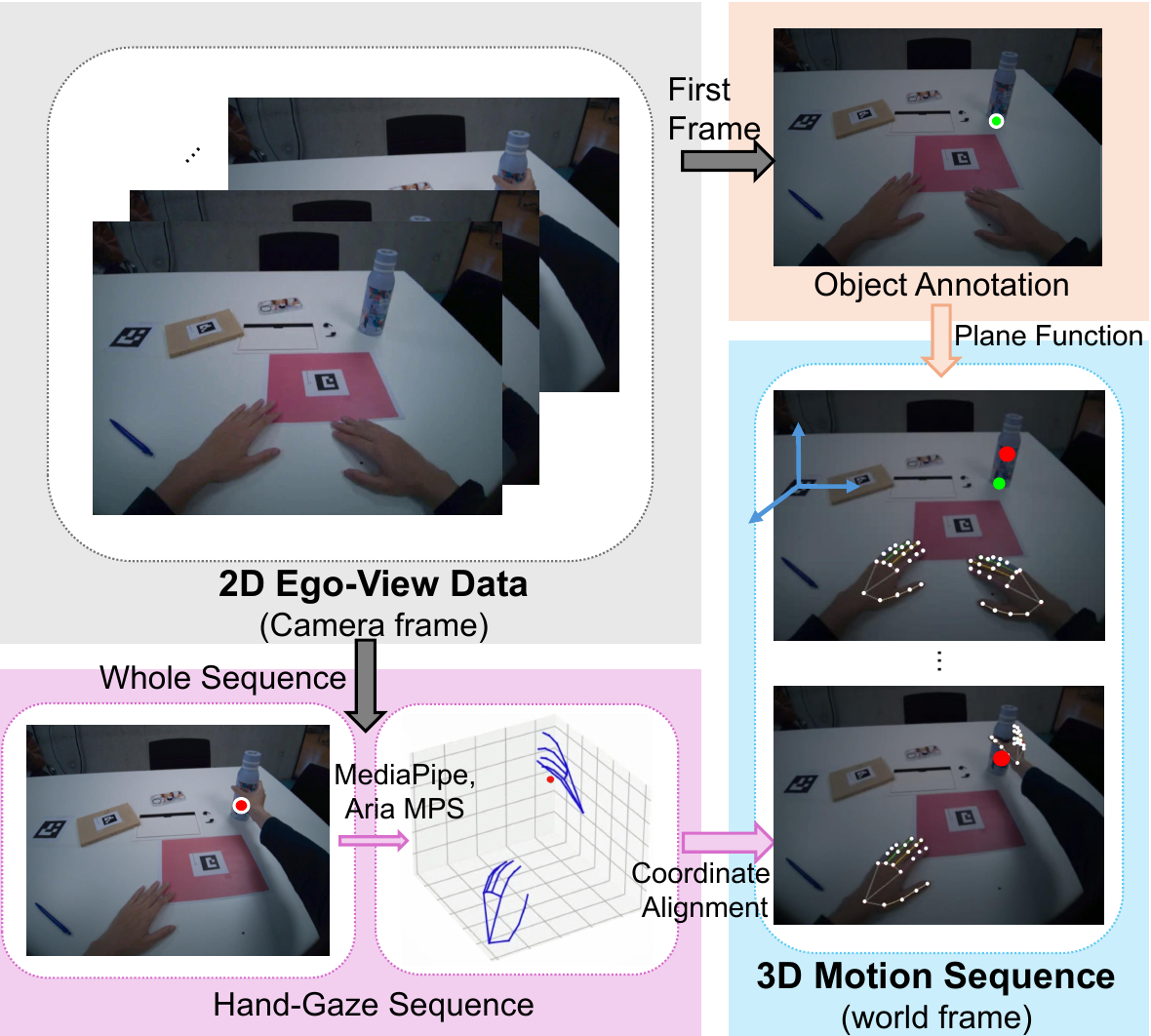}
\caption{\textbf{Data processing pipeline.} This figure illustrates the sequence of steps applied to process egocentric video data for analysis: (a) Raw 2D images are captured from an egocentric-view video. (b) Throughout the entire sequence, the Mediapipe framework and Aria MPS are utilized to extract 3D hand motion, while Aria MPS extracts 3D gaze points. (c) The object representation is manually annotated on the first frame of the video. (d) A world coordinate is employed to integrate the hand-gaze sequence with the object representation into a unified 3D world frame.}
\label{data_collection}
\end{figure}

\begin{figure*}[htbp]
\centering
\subcaptionbox{End-Pose Position Error}{\includegraphics[width=0.9\textwidth]{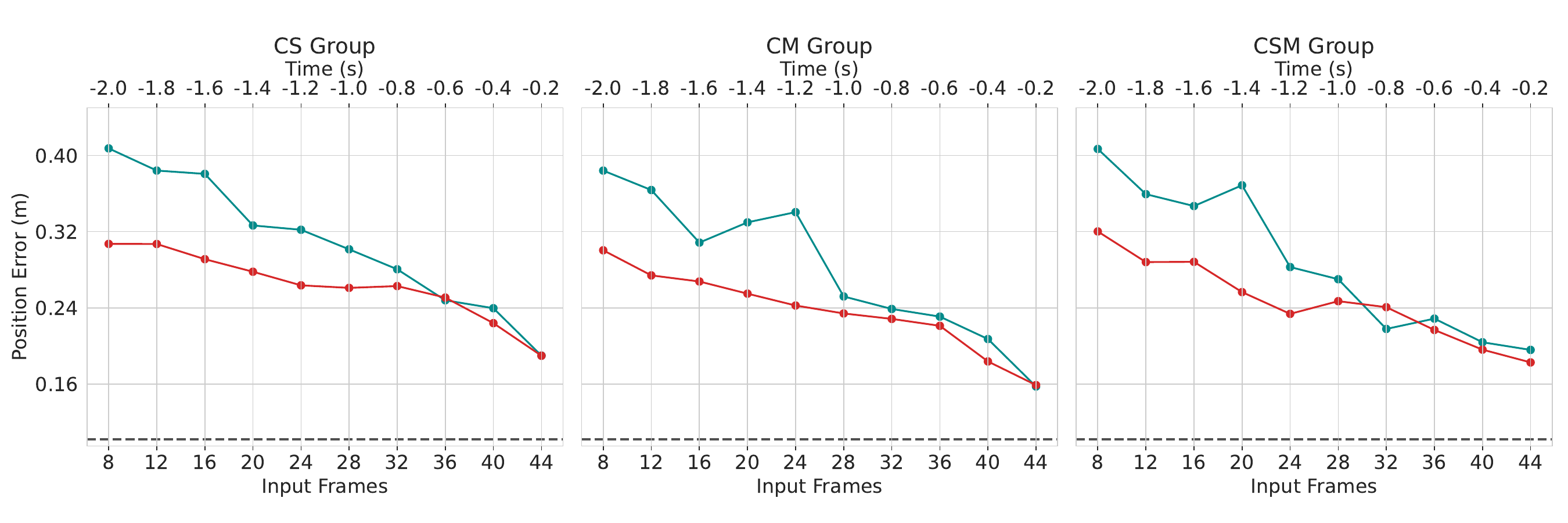}}
\hfill
\subcaptionbox{Average Position Error}{\includegraphics[width=0.9\textwidth]{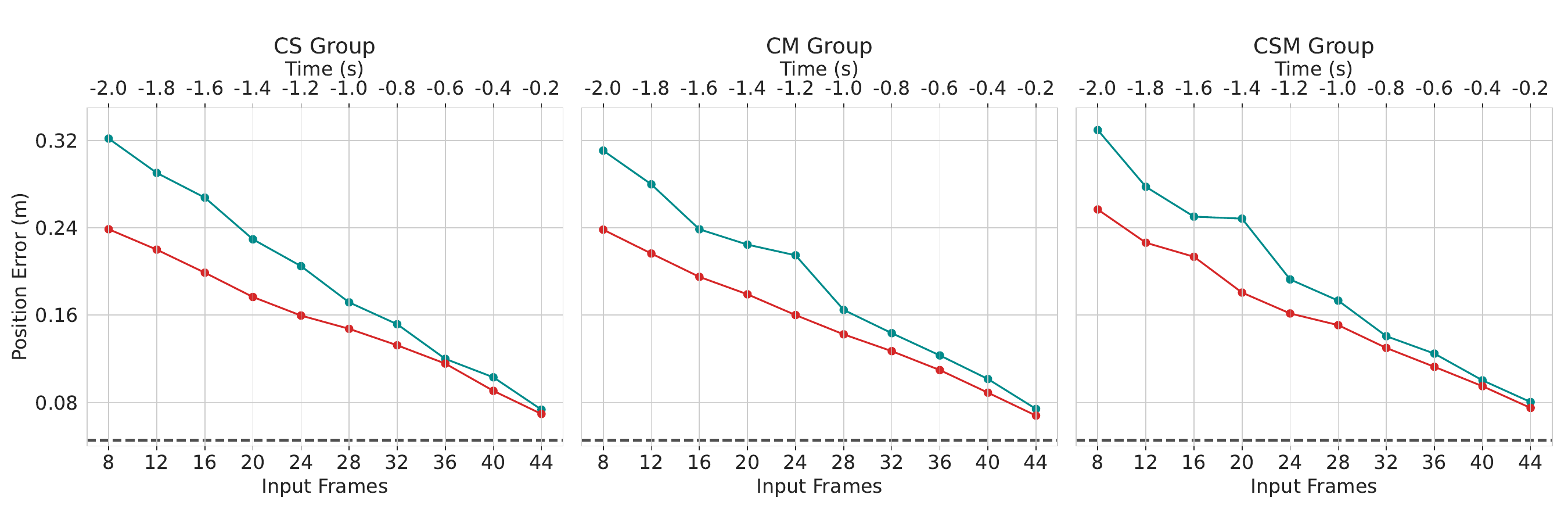}}
\caption{\textbf{Position Errors (in $m$) across Various Input Frames and Time (in $s$).} This figure displays the end-pose (first row) and average (second row) \textit{position} errors within the CS, CM, and CSM groups across different numbers of input frames and time before contact. Red lines represent results with gaze, and green lines represent results without gaze. Gray dashed lines at the bottom represent the position error calculated directly through the encoder and decoder of the hand-pose VQ-VAE.
}
\label{res}
\end{figure*} 


Participants, already equipped with eye-tracking glasses, are seated at a table with their hands placed palms down. An instructor randomly positions a target object on the table for grasping alongside other objects to simulate a real-life scenario. 
Participants are instructed to pick up the object with their preferred hand. Each object is grasped five times by each participant from randomly determined positions, totaling thirty grasping attempts.
With the exception of earphones—which may be placed singly or in pairs, requiring bimanual coordination if paired—all objects can be grasped with one hand. Additionally, a bimanual task—writing on paper—is included and performed once per participant.


\subsubsection{3D Eye-Hand-Object Data Acquisition}
The data acquisition procedure is illustrated in Fig. \ref{data_collection}.
We extract relative 3D hand joints from the 2D video using MediaPipe \cite{lugaresi2019mediapipe}. Subsequently, using the Project Aria Machine Perception Service (MPS), we obtain 3D wrist positions and gaze data from an egocentric viewpoint. By assuming uniform joint-to-wrist lengths across participants, we project these relative 3D joints into a global 3D space based on the known 3D wrist positions. For object positioning, we manually identify their exact locations in the video and map these into 3D space using the plane function defined by the table surface. The 3D eye gaze point is provided by the Aria glasses. Lastly, we synchronize the hand motion, gaze, and object data within a unified coordinate system origin at 1.15 meters to the right and 0.5 meters from the bottom of the table plane.

\subsection{Evaluation Metric and Baseline}

We calculate the Euclidean distance as the position error between the predicted motion and the ground truth as:
\begin{equation}
    e_{\text{position}} = \frac{1}{T} \sum_{t=1}^T \| p_t - \hat{p}_t \|_2,
\end{equation}
where the three dimensional \( p_t \) and \( \hat{p}_t \) are the ground truth and predicted palm positions at each frame \( t \), and \( T \) is the total number of frames.
We compute the position error across all frames and refer to it as \textit{Average Position Error}.
In addition, for the upper extremity rehabilitation robot applications, we are concerned about the prediction of the final grabbing pose given the current input. It has practical usage in giving signals to rehabilitation robots early. To reflect the ability of early prediction, we compute the position error only on the final grabbing pose as \textit{End-Pose Position Error}, which we consider only the last frame ($t=T$), where $p_t=p_T$, $\hat{p}_t=\hat{p}_T$.

We establish a baseline model that only takes hand motion sequences and object embeddings as input, i.e. without the eye-gaze feature. The model architecture remains the same as our proposed model. With this baseline, we want to investigate the effectiveness of the eye gaze feature for the human intention detection task.

\subsection{Cross-Subject and Motion Generalization}
To evaluate the generalization capability of the proposed method in terms of subjects and motions, we design three evaluation settings. 
Two specific actions, ``pick up a book'' and ``write on a piece of paper'', are selected for motion validation that have always been excluded from any training procedure. 
The first evaluation setting is cross-subject (CS), where we perform the five-fold cross-subject evaluation on the 15 subjects from our self-collected dataset. Note that the subjects are different in the training and test sets, while hand actions are the same.
The second evaluation setting is cross-motion (CM), where we train and test on the same groups of subjects while testing only on the ``pick up a book'' and ``write on a piece of paper'' actions that are not presented in the training set. Note that the subjects are the same in the training and test sets, while the hand actions are different.
The third evaluation setting is across both subjects and motions, where we perform the five-fold cross-subject evaluation while only testing on the ``pick up a book'' and ``write on a piece of paper'' actions. Note that both subjects and hand actions are different.

All results reported are derived from this comprehensive cross-validation strategy. 
Position errors are evaluated across a range of input frames from 8 to 44, increasing in increments of four to show the early prediction of the proposed method.

We present the results of end-pose error and average position error in Fig. \ref{res}.
Across all evaluation settings, the position errors demonstrated a decreasing trend as the number of input frames increased for both end-pose and average position errors. Models integrating gaze information generally exhibited lower errors across CS, CM, and CSM settings, although there were exceptions. All three settings exhibited similar position error patterns, indicating that the model generalizes well across different settings. Notably, the disparity in position errors between models with and without gaze became more pronounced with fewer input frames. Gaze-enhanced models showed smaller errors, suggesting the potential of gaze-enhanced models to provide more accurate and immediate corrections in real-time applications where rapid response is crucial.

\begin{table*}[htbp] 
\renewcommand{\arraystretch}{1.3} 
\caption{\textbf{Position error Comparison of End-Pose by Fusion Type.} This table displays the position error (in $m$) for end-pose across different input frames, comparing various fusion types within the CS, CM, and CSM groups.}
\centering
\small
\begin{tabular}{ccc|cccccccccc}
\hline
\multicolumn{2}{c}{Validation Type} & \multicolumn{1}{c|}{Fusion Type} & \multicolumn{10}{c}{Input Frames} \\
\cline{4-13}
\multicolumn{3}{c|}{}               & 8      & 12     & 16     & 20     & 24     & 28     & 32     & 36     & 40     & 44     \\
\hline
\multirow{3}{*}{CS} & \multirow{3}{*}{} 
                    & Linear     & \textbf{0.3071} & \textbf{0.3070} & \textbf{0.2911} & \textbf{0.2779} & \textbf{0.2636} & \textbf{0.2610} & 0.2629 & 0.2508 & \textbf{0.2240} & \textbf{0.1898} \\
                    &                   & Convolution& 0.3320 & 0.3195 & 0.3417 & 0.3044 & 0.2765 & 0.2819 & 0.2705 & 0.2635 & 0.2344 & 0.1954 \\
                    &                   & Summation  & 0.4149 & 0.3265 & 0.3171 & 0.3199 & 0.2829 & 0.2715 & \textbf{0.2611} & \textbf{0.2425} & 0.2260 & 0.1967 \\
\hline
\multirow{3}{*}{CM} & \multirow{3}{*}{} 
                    & Linear     & \textbf{0.3004} & \textbf{0.2741} & \textbf{0.2677} & \textbf{0.2549} & \textbf{0.2425} & \textbf{0.2341} & \textbf{0.2284} & 0.2212 & \textbf{0.1838} & \textbf{0.1590} \\
                    &                   & Convolution& 0.3427 & 0.2988 & 0.3041 & 0.2934 & 0.2810 & 0.2719 & 0.2629 & 0.2475 & 0.2222 & 0.1822 \\
                    &                   & Summation  & 0.3494 & 0.3223 & 0.3146 & 0.2988 & 0.2656 & 0.2561 & 0.2295 & \textbf{0.2007} & 0.1882 & 0.1624 \\
\hline
\multirow{3}{*}{CSM}& \multirow{3}{*}{} 
                    & Linear     & \textbf{0.3202} & \textbf{0.2880} & \textbf{0.2883} & \textbf{0.2566} &\textbf{ 0.2337} & \textbf{0.2470} & 0.2407 & 0.2168 & \textbf{0.1962} & 0.1828 \\
                    &                   & Convolution& 0.3479 & 0.3233 & 0.3309 & 0.3108 & 0.2681 & 0.2665 & 0.2458 & 0.2349 & 0.2210 & 0.1863 \\
                    &                   & Summation  & 0.3315 & 0.3067 & 0.2980 & 0.2805 & 0.2783 & 0.2706 & \textbf{0.2245} & \textbf{0.2146} & 0.2139 & \textbf{0.1713} \\
\hline
\end{tabular}
\label{table:dist}
\end{table*}



The average error throughout the entire grasping process is evaluated similarly to the end-pose error, as depicted in Fig. \ref{res} (b). Across all experimental groups—CS, CM, and CSM—demonstrate consistent trends where models with gaze information outperformed those without. This indicates that gaze information plays a critical role in guiding the movement process. The performance gap between the gaze and no-gaze models becomes more pronounced as the number of input frames decreases, suggesting that gaze information is particularly beneficial in the early stages of input where less historical data is available to aid prediction.

\subsection{Ablation Study on Gaze Fusion Techniques}

To investigate the optimal way of integrating gaze information into the model, we conduct comparisons between our standard linear feature integration method and two simple yet effective gaze-fusion methods: convolutional fusion and direct summation. 
For the convolutional fusion method, we incorporate convolution layers equipped with $1 \times 1$ kernels. In the direct summation method, we first expand the gaze feature to match the dimensionality of the hand motion features before adding them together. 

The results based on the end-pose error across the CS, CM and CSM groups are shown Tab. \ref{table:dist}
From the table, we can see that the linear combination method in our model generally surpasses other methods in reducing position error across the CS, CM, and CSM groups, although there are a few exceptions. Specifically, for shorter input sequences ranging from 4 to 28 frames, the linear combination consistently delivers superior performance compared to other methods. As the length of the input sequences increases, providing a longer historical context, the performance of the other methods becomes comparable. 

\subsection{Visualization of hand motion prediction process}
\begin{figure}[t!]
\centering
\includegraphics[width=0.5\textwidth]{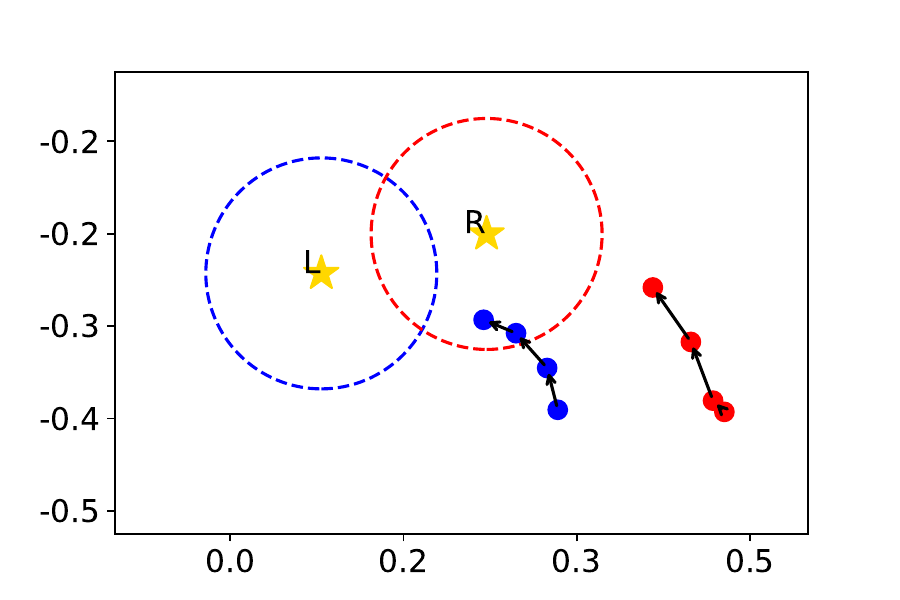}
\caption{\textbf{An example of hand sequential position predictions from top view.} 
The red and blue dots indicate the predicted right and left hand positions respectively, linked by arrows to be the prediction sequences.
The stars mark the final targets for both hands, surrounded by dashed circles denoted as the `target zone'. The axis has a unit of meters.}
\label{visualization}
\end{figure}

In practical applications, the proposed method could be used to predict possible hand positions in the near future, in order to guide the robot to assist individuals in planning the next actions. We take one example from our method and illustrate the progressive process in Fig. \ref{visualization}. 

In the figure, we show the predictions demonstrate a trend of movement of both hands towards the target or the designed target zone. 
The method aims to bring the hand closer to the target, with human intervention taking over once within the target zone. The prediction sequences, visualized by dots linked by arrows, show the hand movements from an initial position progressing towards the target over a period of 0.9 seconds, with updates every 0.3 seconds. 
Given that human movements naturally refine and adjust positioning, the network can utilize these human-corrected positions to further forecast future positions with greater precision. This capability demonstrates the usability of the system in dynamic real-world environments and shows promising research toward robust hand motion prediction.

\section{CONCLUSION}
This study presents an intention detection method that effectively integrates gaze data and egocentric visual cues to predict hand motion sequences, particularly in grasping tasks. By incorporating a VQ-VAE and an auto-regressive generative transformer, our approach not only predicts future hand poses with a high degree of accuracy but also demonstrates robustness against significant noise levels and adaptability to different subjects and objects. These findings highlights the efficacy of our gaze-enhanced model and facilitate its application in real-time interactive environments where rapid and reliable intention detection is critical. 

\section*{ACKNOWLEDGMENT}
We thank Meta for the Aria Glasses in this study.

\bibliographystyle{ieeetr}
\bibliography{ref}


\end{document}